# Analogical Dissimilarity: Definition, Algorithms and Two Experiments in Machine Learning

**Laurent Miclet**                                            LAURENT.MICLET@UNIV-RENNES1.FR
**Sabri Bayoudh**                                            SABRI.BAYOUDH@UNIV-ST-ETIENNE.FR
**Arnaud Delhay**                                            ARNAUD.DELHAY@UNIV-RENNES1.FR
*IRISA/CORDIAL, 6, rue de Kerampont*
*BP 80518 - F-22305 Lannion Cedex, France*

## Abstract

This paper defines the notion of analogical dissimilarity between four objects, with a special focus on objects structured as sequences. Firstly, it studies the case where the four objects have a null analogical dissimilarity, *i.e.* are in analogical proportion. Secondly, when one of these objects is unknown, it gives algorithms to compute it. Thirdly, it tackles the problem of defining analogical dissimilarity, which is a measure of how far four objects are from being in analogical proportion. In particular, when objects are sequences, it gives a definition and an algorithm based on an optimal alignment of the four sequences. It gives also learning algorithms, *i.e.* methods to find the triple of objects in a learning sample which has the least analogical dissimilarity with a given object. Two practical experiments are described: the first is a classification problem on benchmarks of binary and nominal data, the second shows how the generation of sequences by solving analogical equations enables a handwritten character recognition system to rapidly be adapted to a new writer.

## 1. Introduction

Analogy is a way of reasoning that has been studied throughout the history of philosophy and has been widely used in Artificial Intelligence and Linguistics. We focus in this paper on a restricted concept of analogy called 'analogical proportion'.

### 1.1 Analogical Proportion between Four Elements

An *analogical proportion* between four elements $A$, $B$, $C$ and $D$ in the same universe is usually expressed as follows: "$A$ *is to* $B$ *as* $C$ *is to* $D$". Depending on the elements, analogical proportions[1] can have very different meanings. For example, natural language analogical proportions could be: "a crow *is to* a raven *as* a merlin *is to* a peregrine" or "vinegar *is to* wine *as* a sloe *is to* a cherry". They are based on the *semantics* of the words. By contrast, in the formal universe of sequences, analogical proportions such as "abcd *is to* abc *as* abbd *is to* abb" or "g *is to* gt *as* gg *is to* ggt" are *morphological*.

Whether morphological or not, the examples above show the intrinsic ambiguity in defining an analogical proportion. We could as well accept, for other good reasons: "g *is to* gt *as* gg *is to* ggtt" or "vinegar *is to* wine *as* vulgar *is to* wul". Obviously, such ambiguities are inherent in semantic analogies, since they are related to the meaning of words (the concepts are expressed through natural language). Hence, it seems important, as a first step, to focus on formal morphological properties. Moreover, solving such analogies in sequences is an operational problem in several fields of linguistics, such as morphology and syntax, and provides a basis to learning and data mining by analogy in the universe of sequences.

---

1. When there is no ambiguity, we may use 'analogy' for short instead of 'analogical proportion'.

In this paper, we will firstly consider analogical proportions in sets of objects and we will secondly present how they may be transferred to sequences of elements of these sets.

## 1.2 Solving Analogical Equations

When one of the four elements is unknown, an analogical proportion turns into an equation. For instance, on sequences of letters, the analogical proportion "`wolf` *is to* `leaf` *as* `wolves` *is to* $x$" corresponds to the equation $S = \{x \mid$ `wolf` *is to* `leaf` *as* `wolves` *is to* $x\}$. Resolving this equation consists in computing the (possibly empty) set $S$ of sequences $x$ which satisfy the analogy. The sequence `leaves` is an exact semantic and morphological solution. We shall see that, however, it is not straightforward to design an algorithm able to solve this kind of equation, in particular when looking for an approximate solution if necessary.

Solving analogical equations on sequences is useful for linguistic analysis tasks and has been applied (with empirical resolution techniques, or in simple cases) mainly to lexical analysis tasks. For example, Yvon (1999) presents an analogical approach to the grapheme-to-phoneme conversion, for text-to-speech synthesis purposes. More generally, the resolution of analogical equations can also be seen as a basic component of *learning by analogy* systems, which are part of the *lazy learning* techniques (Daelemans, 1996).

## 1.3 Using Analogical Proportions in Machine Learning

Let $\mathcal{S} = \{(x, u(x))\}$ be a finite set of training examples, where $x$ is the description of an example ($x$ may be a sequence or a vector in $\mathbb{R}^n$, for instance) and $u(x)$ its label in a finite set. Given the description $y$ of a new pattern, we would like to assign to $y$ a label $u(y)$, based only from the knowledge of $\mathcal{S}$. This is the problem of inductive learning of a classification rule from examples, which consists in finding the value of $u$ at point $y$ (Mitchell, 1997). The nearest neighbor method, which is the most popular lazy learning technique, simply finds in $\mathcal{S}$ the description $x^\star$ which minimizes some distance to $y$ and hypothesizes $u(x^\star)$, the label of $x^\star$, for the label of $y$.

Moving one step further, learning from analogical proportions consists in searching in $\mathcal{S}$ for a triple $(x^\star, z^\star, t^\star)$ such that "$x^\star$ *is to* $z^\star$ *as* $t^\star$ *is to* $y$" and predicts for $y$ the label $\hat{u}(y)$ which is solution of the equation "$u(x^\star)$ *is to* $u(z^\star)$ *as* $u(t^\star)$ *is to* $\hat{u}(y)$". If more than one triple is found, a voting procedure can be used. Such a learning technique is based on the resolution of analogical equations. Pirrelli and Yvon (1999) discuss the relevance of such a learning procedure for various linguistic analysis tasks. It is important to notice that $y$ and $u(y)$ are in different domains: for example, in the simple case of learning a classification rule, $y$ may be a sequence whereas $u$ is a class label.

The next step in learning by analogical proportions is, given $y$, to find a triple $(x^\star, z^\star, t^\star)$ in $\mathcal{S}$ such that "$x^\star$ *is to* $z^\star$ *as* $t^\star$ *is to* $y$" holds *almost* true, or, when a closeness measure is defined, the triple which is the closest to $y$ in term of analogical proportion. We study in this article how to quantify this measure, in order to provide a more flexible method of learning by analogy.

### 1.4 Related Work

This paper is related with several domains of artificial intelligence. Obviously, the first one is that of reasoning by analogy. Much work has been done on this subject from a cognitive science point of view, which had led to computational models of reasoning by analogy: see for example, the classical paper (Falkenhainer, Forbus, & Gentner, 1989), the book (Gentner, Holyoak, & Kokinov, 2001) and the recent survey (Holyoak, 2005). Usually, these works use the notion of *transfer*, which is not within the scope of this article. It means that some knowledge on solving a problem in a domain is transported to another domain. Since we work on four objects that are in the same space, we implicitly ignore the notion of transfer between *different* domains. Technically speaking, this restriction allows us to use an axiom called 'exchange of the means' to define an analogical proportion (see Definition 2.1). However, we share with these works the following idea: there may be a similar relation between two couples of structured objects even if the objects are apparently quite different. We are interested in giving a formal and algorithmic definition of such a relation.

Our work also aims to define some supervised machine learning process (Mitchell, 1997; Cornuéjols & Miclet, 2002), in the spirit of *lazy* learning. We do not seek to extract a model from the learning data, but merely conclude what is the class, or more generally the supervision, of a new object by inspecting (a part of) the learning data. Usually, lazy learning, like the $k$-nearest neighbors technique, makes use of unstructured objects, such as vectors. Since distance measures can be also defined on strings, trees and even graphs, this technique has also been used on structured objects, in the framework of structural pattern recognition (see for example the work of Bunke & Caelli, 2004; Blin & Miclet, 2000; Basu, Bunke, & Del Bimbo, 2005). We extend here the search of the nearest neighbor in the learning set to that of the best triple (when combined with the new object, it is the closest to make an analogical proportion). This requires defining what is an analogical proportion on structured objects, like sequences, but also to give a definition of how far a 4-tuple of objects is from being in analogy (that we call analogical dissimilarity).

Learning by analogy on sequences has already being studied, in a more restricted manner, on linguistic data (Yvon, 1997, 1999; Itkonen & Haukioja, 1997). Reasoning and learning by analogy has proven useful in tasks like grapheme to phoneme conversion, morphology and translation. Sequences of letters and/or of phonemes are a natural application to our work, but we are also interested in other type of data, structured as sequences or trees, such as prosodic representations for speech synthesis, biochemical sequences, online handwriting recognition, etc.

Analogical proportions between four structured objects of the same universe, mainly strings, have been studied with a mathematical and algorithmic approach, like ours, by Mitchell (1993) and Hofstadter et al. (1994), Dastani et al. (2003), Schmid et al. (2003). To the best of our knowledge our proposition is original: to give a formal definition of what can be an analogical dissimilarity between four objects, in particular between sequences, and to produce algorithms that enable the efficient use of this concept in machine learning practical problems. We have already discussed how to compute exact analogical proportions between sequences in the paper by Yvon et al. (2004) and given a preliminary attempt to compute analogical dissimilarity between sequences in the paper by Delhay and Miclet (2004). Excerpts of the present article have been presented in conferences (Bayoudh, Miclet, & Delhay, 2007a; Bayoudh, Mouchère, Miclet, & Anquetil, 2007b).

To connect with another field of A.I., let us quote Aamodt and Plaza (1994) about the use of the term 'analogy' in Case-Based Reasoning (CBR): 'Analogy-based reasoning: This term is sometimes used, as a synonym to case-based reasoning, to describe the typical case-based approach.

However, it is also often used to characterize methods that solve new problems based on past cases from a different domain, while typical case-based methods focus on indexing and matching strategies for single-domain cases.' According to these authors, who use the word 'analogy' in its broader meaning, typical CBR deals with single domain problems, as analogical proportions also do. In that sense, our study could be seen as a particular case of CBR, as applied in this paper to supervised learning of classification rules.

### 1.5 Organization of the Paper

This paper is organized in six sections. After this introduction, we present in section 2 the general principles which govern the definition of an analogical proportion between four objects in the same set and we define what is an analogical equation in a set. We apply these definitions in $\mathbb{R}^n$ and $\{0, 1\}^n$. Finally, this section defines analogical proportion between four sequences on an alphabet in which an analogy is defined, using an optimal alignment method between the four sequences.

Sections 3 introduces the new concept of *analogical dissimilarity* ($AD$) between four objects, by measuring in some way how much these objects are in analogy. In particular, it must be equivalent to say that four objects are in analogy and that their analogical dissimilarity is null. Then we extend it to sequences. The end of this section gives two algorithms: SEQUANA4 computes the value of $AD$ between four sequences and SOLVANA solves analogical equations in a generalized manner: it can produce approximate solutions (*i.e.* of strictly positive $AD$).

Section 4 begins to explore the use of the concept of analogical dissimilarity in supervised machine learning. We give an algorithm (FADANA) for the fast search of the $k$-best analogical 3-tuples in the learning set.

Section 5 presents two applications of these concepts and algorithms on real problems. We firstly apply FADANA to objects described by binary and nominal features. Experiments are conducted on classical benchmarks and favorably compared with standard classification techniques. Secondly, we make use of SOLVANA to produce new examples in a handwritten recognition system. This allows training a classifier from a very small number of learning patterns.

The last section presents work to be done, particularly in discussing more real world application of learning by analogy, especially in the universe of sequences.

## 2. Analogical Proportions and Equations

In this section, we give a formal definition of the *analogical proportion* between four objects and explain what is to solve an *analogical equation*. Instanciations of the general definitions are given when the objects are either finite sets (or equivalently binary vectors), or vectors of real numbers or sequences on finite alphabets.

### 2.1 The Axioms of Analogical Proportion

The meaning of an analogical proportion $A : B :: C : D$ between four objects in a set $X$ depends on the nature of $X$, in which the 'is to' and the 'as' relations have to be defined. However, general properties can be required, according to the usual meaning of the word 'analogy' in philosophy and linguistics. According to Lepage (2003) three basic axioms can be given:

**Definition 2.1 (Analogical proportion)** *An analogical proportion on a set $X$ is a relation on $X^4$, i.e. a subset $\mathcal{A} \subset X^4$. When $(A, B, C, D) \in \mathcal{A}$, the four elements $A$, $B$, $C$ and $D$ are said* to be in analogical proportion, *and we write: 'the analogical proportion* $A : B :: C : D$ *holds true', or simply* $A : B :: C : D$ *, which reads 'A is to B as C is to D'. For every 4-tuple in analogical proportion, the following equivalences must hold true:*

Symmetry of the 'as' relation*:* $\quad A : B :: C : D \;\Leftrightarrow\; C : D :: A : B$

Exchange of the means*:* $\quad A : B :: C : D \;\Leftrightarrow\; A : C :: B : D$

*The third axiom (*determinism*) requires that one of the two following implications holds true (the other being a consequence):*

$$A : A :: B : x \quad \Rightarrow \quad x = B$$
$$A : B :: A : x \quad \Rightarrow \quad x = B$$

According to the first two axioms, five other formulations are equivalent to the canonical form $A : B :: C : D$ :

$$B : A :: D : C \qquad D : B :: C : A \qquad C : A :: D : B$$
$$D : C :: B : A \qquad \text{and} \qquad B : D :: A : C$$

Consequently, there are only three different possible analogical proportions between four objects, with the canonical forms:

$$A : B :: C : D \qquad A : C :: D : B \qquad A : D :: B : C$$

## 2.2 Analogical Equations

To solve an analogical equation consists in finding the fourth term of an analogical proportion, the first three being known.

**Definition 2.2 (Analogical equation)** *$D$ is a solution of the analogical equation* $A : B :: C : x$ *if and only if* $A : B :: C : D$ *.*

We already know from previous sections that, depending on the nature of the objects and the definition of analogy, an analogical equation may have either no solution or a unique solution or several solutions. We study in the sequel how to solve analogical equations in different sets.

## 2.3 Analogical Proportion between Finite Sets and Binary Objects

When the 'as' relation is the equality between sets, Lepage has given a definition of an analogical proportion between sets coherent with the axioms. This will be useful in section 2.3.2 in which objects are described by sets of binary features.

### 2.3.1 AN ANALOGICAL PROPORTION IN FINITE SETS

**Definition 2.3 (Analogical proportion between finite sets)** *Four sets $A$, $B$, $C$ and $D$ are in analogical proportion* $A : B :: C : D$ *if and only if $A$ can be transformed into $B$, and $C$ into $D$, by adding and subtracting the same elements to $A$ and $C$.*

This is the case, for example, of the four sets: $A = \{t_1, t_2, t_3, t_4, \}$, $B = \{t_1, t_2, t_3, t_5\}$ and $C = \{t_1, t_4, t_6, t_7\}$, $D = \{t_1, t_5, t_6, t_7\}$, where $t_4$ has been taken off from, and $t_5$ has been added to $A$ and $C$, giving $B$ and $D$.

### 2.3.2 SOLVING ANALOGICAL EQUATIONS IN FINITE SETS

Considering analogy in sets, Lepage (2003) has shown the following theorem, with respect to the axioms of analogy (section 2.1):

**Theorem 2.4 (Solution of an analogical equation in sets)** *Let $A$, $B$ and $C$ be three sets. The analogical equation $A : B :: C : D$ where $D$ is the unknown has a solution if and only if the following conditions hold true:*

$$A \subseteq B \cup C \ \text{ and } \ A \supseteq B \cap C$$

*The solution is then unique, given by:*

$$D = ((B \cup C) \backslash A) \cup (B \cap C)$$

### 2.3.3 ANALOGICAL PROPORTIONS IN $\{0, 1\}^n$

Let now $X$ be the set $\{0, 1\}^n$. For each $x \in X$ and each $i \in [1, n]$, $f_i(x) = 1$ (resp. $f_i(x) = 0$) means that the binary feature $f_i$ takes the value $TRUE$ (resp. $FALSE$) on the object $x$.

Let $A : B :: C : D$ be an analogical equation. For each feature $f_i$, there are only eight different possibilities of values on $A$, $B$ and $C$. We can derive the solutions from the definition and properties of analogy on sets, with the two following principles:

- Each feature $f_i(D)$ can be computed independently.

- The following table gives the solution $f_i(D)$:

| $f_i(A)$ | 0 | 0 | 0 | 0 | 1 | 1 | 1 | 1 |
|---|---|---|---|---|---|---|---|---|
| $f_i(B)$ | 0 | 0 | 1 | 1 | 0 | 0 | 1 | 1 |
| $f_i(C)$ | 0 | 1 | 0 | 1 | 0 | 1 | 0 | 1 |
| $f_i(D)$ | 0 | 1 | 1 | ? | ? | 0 | 0 | 1 |

In two cases among the eight, $f_i(D)$ does not exists. This derives from the defining of $X$ by binary features, which is equivalent to defining $X$ as a finite set. Theorem 2.4 imposes conditions on the resolution of analogical equations on finite sets, which results in the fact that two binary analogical equations have no solution.

## 2.4 Analogical Proportion in $\mathbb{R}^n$

### 2.4.1 DEFINITION

Let $O$ be the origin of $\mathbb{R}^n$. Let $a = (a_1, a_2, \dots, a_n)^{\mathsf{T}}$ be a vector of $\mathbb{R}^n$, as defined by its $n$ coordinates. Let $a$, $b$, $c$ and $d$ be four vectors of $\mathbb{R}^n$. The interpretation of an analogical proportion $a : b :: c : d$ is usually that $a$, $b$, $c$, $d$ are the corners of a parallelogram, $a$ and $d$ being opposite corners (see Figure 1).

**Definition 2.5 (Analogical proportion in $\mathbb{R}^n$)** *Four elements of $\mathbb{R}^n$ are in the analogical proportion ($a : b :: c : d$) if and only if they form a parallelogram, that is when $\overrightarrow{Oa} + \overrightarrow{Od} = \overrightarrow{Ob} + \overrightarrow{Oc}$ or equivalently $\overrightarrow{ab} = \overrightarrow{cd}$ or equivalently $\overrightarrow{ac} = \overrightarrow{bd}$*

It is straightforward that the axioms of analogy, given in section 2.1 are verified by this definition.

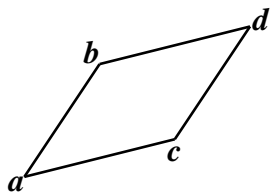

Figure 1: Analogical parallelogram in $\mathbb{R}^n$.

### 2.4.2 SOLVING ANALOGICAL EQUATIONS IN $\mathbb{R}^n$

Solving the analogical equation $a : b :: c : x$, where $a$, $b$ and $c$ are vectors of $\mathbb{R}^n$ and $x$ is the unknown derives directly from the definition of analogy in vector spaces: the four vectors must form a parallelogram. There is always one and only one solution given by the equation:

$$\overrightarrow{Ox} = \overrightarrow{Ob} + \overrightarrow{Oc} - \overrightarrow{Oa}$$

## 2.5 Analogical Proportion between Sequences

### 2.5.1 NOTATIONS

A *sequence*[2] is a finite series of symbols from a finite alphabet $\Sigma$. The set of all sequences is denoted $\Sigma^\star$. For $x$, $y$ in $\Sigma^\star$, $xy$ denotes the concatenation of $x$ and $y$. We also denote $|x| = n$ the length of $x$, and we write $x$ as $x = x_1 \ldots x_{|x|}$ or $x = x[1] \ldots x[n]$, with $x_i$ or $x[i] \in \Sigma$. We denote $\epsilon$ the empty word, of null length, and $\Sigma^+ = \Sigma^\star \backslash \{\epsilon\}$.

A *factor* (or subword) $f$ of a sequence $x$ is a sequence in $\Sigma^\star$ such that there exists two sequences $u$ and $v$ in $\Sigma^\star$ with: $x = ufv$. For example, $abb$ and $bbac$ are factors of $abbacbbaba$.

A *subsequence* of a sequence $x = x_1 \ldots x_{|x|}$ is composed of the letters of $x$ with the indices $i_1 \ldots i_k$, such that $i_1 < i_2 \ldots < i_k$. For example, $ca$ and $aaa$ are two subsequences of $abbacbaba$.

### 2.5.2 DEFINITION

Let $\Sigma$ be an alphabet. We add a new letter to $\Sigma$, that we denote $\smallsmile$, giving the augmented alphabet $\Sigma'$. The interpretation of this new letter is simply that of an 'empty' symbol, that we will need in subsequent sections.

**Definition 2.6 (Semantic equivalence)** *Let $x$ be a sequence of $\Sigma^\star$ and $y$ a sequence of $\Sigma'^\star$. $x$ and $y$ are* semantically equivalent *if the subsequence of $y$ composed of letters of $\Sigma$ is $x$. We denote this relation by $\equiv$.*

For example, $ab \smallsmile a \smallsmile a \equiv abaa$.

Let us assume that there is an analogy in $\Sigma'$, i.e. that for every 4-tuple $a$, $b$, $c$, $d$ of letters of $\Sigma'$, the relation $a : b :: c : d$ is defined as being either $TRUE$ or $FALSE$.

**Definition 2.7 (Alignment between two sequences)** *An* alignment *between two sequences $x, y \in \Sigma^\star$, of lengths $m$ and $n$, is a word $z$ on the alphabet $(\Sigma') \times (\Sigma') \backslash \{(\smallsmile, \smallsmile)\}$ whose first projection is semantically equivalent to $x$ and whose second projection is semantically equivalent to $y$.*

---

2. More classically in language theory, a *word* or a *sentence*.

Informally, an alignment represents a one-to-one letter matching between the two sequences, in which some letters ⌣ may be inserted. The matching (⌣, ⌣) is not permitted. An alignment can be presented as an array of two rows, one for $x$ and one for $y$, each word completed with some ⌣, resulting in two words of $\Sigma'$ having the same length.

For instance, here is an *alignment* between $x = abgef$ and $y = acde$ :

$$
\begin{array}{ccccccc}
x' & = & a & b & \smile & g & e & f \\
  &   & | & | & | & | & | & | \\
y' & = & a & c & d & \smile & e & \smile
\end{array}
$$

We can extend this definition to that of an alignment between four sequences.

**Definition 2.8 (Alignment between four sequences)** *An* alignment *between four sequences* $u, v, w, x \in \Sigma^\star$, *is a word $z$ on the alphabet* $(\Sigma \cup \{\smile\})^4 \backslash \{(\smile, \smile, \smile, \smile)\}$ *whose projection on the first, the second, the third and the fourth component is respectively semantically equivalent to* $u$, $v$, $w$ *and* $x$.

The following definition uses alignments between four sequences.

**Definition 2.9 (Analogical proportion between sequences)** *Let $u$, $v$, $w$ and $x$ be four sequences on $\Sigma^\star$, on which an analogy is defined. We say that $u$, $v$, $w$ and $x$ are in* analogical proportion *in $\Sigma^\star$ if there exists four sequences $u'$, $v'$, $w'$ and $x'$ of same length $n$ in $\Sigma'$, with the following properties:*

1. *$u' \equiv u$, $v' \equiv v$, $w' \equiv w$ and $x' \equiv x$.*

2. *$\forall i \in [1, n]$ the analogies $u_i' : v_i' :: w_i' : x_i'$ hold true in $\Sigma'$.*

One has to note that Lepage (2001) and Stroppa and Yvon (2004) have already proposed a definition of an analogical proportion between sequences with applications to linguistic data. Basically, the difference is that they accept only trivial analogies in the alphabet (such as $a : b :: a : b$ or $a : \smile :: a : \smile$).

For example, let $\Sigma' = \{a, b, \alpha, \beta, B, C, \smile\}$ with the non trivial analogies $a : b :: A : B$, $a : \alpha :: b : \beta$ and $A : \alpha :: B : \beta$. The following alignment between the four sequences $aBA$, $\alpha bBA$, $ba$ and $\beta ba$ is an analogical proportion on $\Sigma^\star$:

$$
\begin{array}{cccc}
a & \smile & B & A \\
\alpha & b & B & A \\
b & \smile & a & \smile \\
\beta & b & a & \smile
\end{array}
$$

## 3. Analogical Dissimilarity

### 3.1 Motivation

In this section, we are interested in defining what could be a relaxed analogy, which linguistic expression would be '$a$ is to $b$ *almost as* $c$ is to $d$'. To remain coherent with our previous definitions, we measure the term 'almost' by some positive real value, equal to 0 when the analogy stands true, and increasing when the four objects are less likely to be in analogy. We also want this value, that we call 'analogical dissimilarity' (AD), to have good properties with respect to the analogy. We want it to be symmetrical, to stay unchanged when we permute the mean terms of the analogy and finally to respect some triangle inequality. These requirements will allow us, in section 4, to generalize a classical fast nearest neighbor search algorithm and to exhibit an algorithmic learning process which principle is to extract, from a learning set, the 3-tuple of objects that has the least AD when combined with another unknown object. This lazy learning technique is a therefore a generalization of the nearest neighbor method.

We firstly study the definition of the analogical dissimilarity on the same structured sets as in the previous sections, and secondly extend it to sequences.

## 3.2 A Definition in $\{0,1\}^n$

**Definition 3.1 (Analogical dissimilarity in $\{0,1\}$)** *The analogical dissimilarity between four binary values is given by the following table:*

| $u$ | 0 | 0 | 0 | 0 | 0 | 0 | 0 | 0 | 1 | 1 | 1 | 1 | 1 | 1 | 1 | 1 |
|---|---|---|---|---|---|---|---|---|---|---|---|---|---|---|---|---|
| $v$ | 0 | 0 | 0 | 0 | 1 | 1 | 1 | 1 | 0 | 0 | 0 | 0 | 1 | 1 | 1 | 1 |
| $w$ | 0 | 0 | 1 | 1 | 0 | 0 | 1 | 1 | 0 | 0 | 1 | 1 | 0 | 0 | 1 | 1 |
| $x$ | 0 | 1 | 0 | 1 | 0 | 1 | 0 | 1 | 0 | 1 | 0 | 1 | 0 | 1 | 0 | 1 |
| $AD(u,v,w,t)$ | 0 | 1 | 1 | 0 | 1 | 0 | 2 | 1 | 1 | 2 | 0 | 1 | 0 | 1 | 1 | 0 |

In other words, the $AD$ between four binary values is the minimal number of bits that have to be switched in order to produce an analogical proportion. It can be seen as an extension of the edit distance in four dimensions which supports the coherence with analogy.

**Definition 3.2 (Analogical dissimilarity in $\{0,1\}^n$)** *The analogical dissimilarity $AD(u,v,w,t)$ between four objects $u$, $v$, $w$ and $t$ of a finite set $X$ defined by binary features is the sum of the values of the analogical dissimilarities between the features.*

### 3.2.1 PROPERTIES

With this definition, the analogical dissimilarity has the following properties:

**Property 3.1 (Properties of $AD$ in $\{0,1\}^n$)**

**Coherence with analogy.**
   $(AD(u,v,w,x) = 0) \Leftrightarrow u : v :: w : x$

**Symmetry for 'as'.** $AD(u,v,w,x) = AD(w,x,u,v)$

**Exchange of medians.** $AD(u,v,w,x) = AD(u,w,v,x)$

**Triangle inequality.** $AD(u,v,z,t) \leq AD(u,v,w,x) + AD(w,x,z,t)$

**Asymmetry for 'is to'.** *In general:* $AD(u,v,w,x) \neq AD(v,u,w,x)$

The first properties are quite straightforward from the definition. The demonstration of the third one is simple as well. If the property

$$AD(f_i(u), f_i(v), f_i(z), f_i(t)) \leq AD(f_i(u), f_i(v), f_i(w), f_i(x))$$
$$+ AD(f_i(w), f_i(x), f_i(z), f_i(t))$$

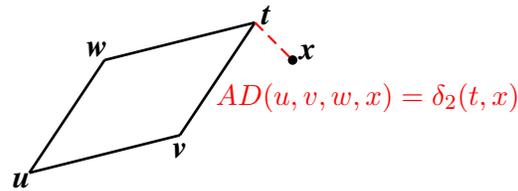

Figure 2: Analogical dissimilarity in vector spaces with distance $\delta_2$.

holds true for every 6-tuple of elements and every feature $f_i$, then property (4) is true. The demonstration being done by examining all possible cases: it is impossible to find 6 binary features $a$, $b$, $c$, $d$, $e$, $f$ such that $AD(a, b, e, f) = 2$ and $AD(a, b, c, d) + AD(c, d, e, f) < 2$. More precisely, if $AD(a, b, e, f) = 2$, $AD(a, b, c, d) + AD(c, d, e, f)$ is also equal to 2 for all the four values that $(c, d)$ can take.

### 3.3 Analogical Dissimilarity in $\mathbb{R}^n$

The analogical dissimilarity between four vectors must reflect in some way how far they are from constructing a parallelogram. Four vectors $u$, $v$, $w$ and $x$ are in analogical proportion (*i.e.*, form a parallelogram) with opposite sides $\overrightarrow{uv}$ and $\overrightarrow{wx}$ if and only if $\overrightarrow{Ou} + \overrightarrow{Ox} = \overrightarrow{Ov} + \overrightarrow{Ow}$, or equivalently $u + x = v + w$, we have chosen the following definition (see Figure 2):

**Definition 3.3 (Analogical dissimilarity between vectors)** *The analogical dissimilarity between four vectors $u$, $v$, $w$ and $x$ of $\mathbb{R}^n$ in which is defined the norm $\| \ \|_p$ and the corresponding distance $\delta_p$ is given by the real positive value $AD(u, v, w, x) = \delta_p(u + x, v + w) = \|(u + x) - (v + w)\|_p$. It is also equal to $\delta_p(t, x)$, where $t$ is the solution of the analogical equation $u : v :: w : t$.*

**Property 3.2 (Properties of $AD$ between vectors)** *This definition of analogical dissimilarity in $\mathbb{R}^n$ guarantees that the following properties hold true: coherence with analogy, symmetry for 'as', exchange of medians, triangle inequality and asymmetry for 'is to'.*

The first two properties are quite straightforward from the definition. Since $\| \ \|_p$ is a norm, it respects the triangle inequality which involves the third property:

$$AD(u, v, z, t) \leq AD(u, v, w, x) + AD(w, x, z, t)$$

### 3.4 Analogical Dissimilarity between Sequences

We present in the following a definition and two algorithms. Firstly, we extend the notion of analogical dissimilarity to sequences. The first algorithm, called SEQUANA4, computes the analogical dissimilarity between four sequences of $\Sigma^\star$. The second one, called SOLVANA, given an analogical equation on sequences, produces the Directed Acyclic Graph (DAG) of all the solutions. If there is no solution, it gives the DAG of all the sentences that have the least analogical dissimilarity when associated with the three known sentences of the equation.

These two algorithms are quite general, since they make no particular assumption on the alphabet of the sequences. This alphabet $\Sigma$ is simply augmented to $\Sigma' = \Sigma \cup \{\frown\}$ to produce alignments as described in section 2.5. The analogical dissimilarity on $\Sigma'$ must be such that: $AD(\frown, \frown, a, a) = 0$, and $AD(\frown, a, b, c) > 0$ for every $a, b, c \in \Sigma$, but no more constraint is required.

### 3.4.1 DEFINITION

Let $\Sigma$ be a set on which is defined an analogical dissimilarity $AD$. We augment it to $\Sigma'$ by adding the special symbol $\frown$. We assume now that there is an analogical dissimilarity $AD$ on $\Sigma'$.

**Definition 3.4 (Analogical dissimilarity between four sequences)** *The cost of an alignment between four sequences is the sum of the analogical dissimilarities between the 4-tuples of letters given by the alignment.*

*The analogical dissimilarity $AD(u, v, w, x)$ between four sequences in $\Sigma^\star$ is the cost of an alignment of minimal cost of the four sequences.*

This definition ensures that the following properties hold true: coherence with analogy, symmetry for 'as', exchange of medians and asymmetry for 'is to'[3].

Depending on what are we looking for, many methods have been developed for multiples alignment in bio-informatics (Needleman & Wunsch, 1970; Smith & Waterman, 1981) :

1. For structure or functional similarity like in protein modelization, pattern identification or structure prediction in DNA, methods using simultaneous alignment like MSA (Wang & Jiang, 1994) or DCA (Dress, Füllen, & Perrey, 1995), or iterative alignment like MUSCLE (Edgar, 2004) are the best.

2. For Evolutionary similarity like in phylogenic classification, methods using progressive alignment and tree structure, like ClustalW (Thompson, Higgins, & Gibson, 1994), are the most fitted.

However, all of these alignment methods (global or local) are heuristic algorithms to overcome the problem of time and space complexity introduced first by the length of sequences and second by the number of the sequences to align. In our generation problem neither the sequence length which is around 30 characters nor the number of sequences to align which is always four in analogy need a heuristic alignment to speed up the algorithm. But techniques used in bio-informatics to compute automatically the substitution matrix could be very helpful and interesting in handwritten characters recognition. Introducing Gap (Gep, Gop) penalties like in DNA or protein sequences should also be an interesting idea to explore.

## 3.5 Computing the Analogical Dissimilarity between Four Sequences: the SEQUANA4 Algorithm

We compute $AD(u, v, w, x)$ with a dynamic programming algorithm, called SEQUANA4, that progresses in synchronicity in the four sequences to build an optimal alignment.

---

3. With this definition of AD, the 'triangle inequality' property is not always true on sequences.

The input of this algorithm is the augmented alphabet $\Sigma'$ on which there an analogical dissimilarity $AD(a, b, c, d)$. The output is the analogical dissimilarity between four sentences of $\Sigma^\star$, namely $AD(u, v, w, x)$.

We give below the basics formulas of the recurrence. When implementing the computation, one has to check the correct progression of the indexes $i$, $j$, $k$ and $l$.

**Initialisation**

$C_{w_0 x_0}^{u_0 v_0} \leftarrow 0$ ;

**for** $i = 1, |u|$ **do** $C_{w_0 x_0}^{u_i v_0} \leftarrow C_{w_0 x_0}^{u_{i-1} v_0} + AD(u_i, \frown, \frown, \frown)$ **done** ;

**for** $j = 1, |v|$ **do** $C_{w_0 x_0}^{u_0 v_j} \leftarrow C_{w_0 x_0}^{u_0 v_{j-1}} + AD(\frown, v_j, \frown, \frown)$ **done** ;

**for** $k = 1, |w|$ **do** $C_{w_k x_0}^{u_0 v_0} \leftarrow C_{w_{k-1} x_0}^{u_0 v_0} + AD(\frown, \frown, w_k, \frown)$ **done** ;

**for** $l = 1, |x|$ **do** $C_{w_0 x_l}^{u_0 v_0} \leftarrow C_{w_0 x_{l-1}}^{u_0 v_0} + AD(\frown, \frown, \frown, x_l)$ **done** ;

**Recurrence**

$$
C_{w_k x_l}^{u_i v_j} = Min \begin{cases}
C_{w_{k-1} x_{l-1}}^{u_{i-1} v_{j-1}} + AD(u_i, v_j, w_k, x_l) & [i \leftarrow i+1; j \leftarrow j+1; k \leftarrow k+1; l \leftarrow l+1] \\
C_{w_{k-1} x_l}^{u_{i-1} v_{j-1}} + AD(u_i, v_j, w_k, \frown) & [i \leftarrow i+1; j \leftarrow j+1; k \leftarrow k+1] \\
C_{w_k x_{l-1}}^{u_{i-1} v_{j-1}} + AD(u_i, v_j, \frown, x_l) & [i \leftarrow i+1; j \leftarrow j+1; l \leftarrow l+1] \\
C_{w_k x_l}^{u_{i-1} v_{j-1}} + AD(u_i, v_j, \frown, \frown) & [i \leftarrow i+1; j \leftarrow j+1] \\
C_{w_{k-1} x_{l-1}}^{u_i v_{j-1}} + AD(\frown, v_j, w_k, x_l) & [j \leftarrow j+1; k \leftarrow k+1; l \leftarrow l+1] \\
C_{w_k x_{l-1}}^{u_i v_{j-1}} + AD(\frown, v_j, \frown, x_l) & [j \leftarrow j+1; l \leftarrow l+1] \\
C_{w_{k-1} x_l}^{u_i v_{j-1}} + AD(\frown, v_j, w_k, \frown) & [i \leftarrow i+1; k \leftarrow k+1] \\
C_{w_k x_l}^{u_i v_{j-1}} + AD(\frown, v_j, \frown, \frown) & [j \leftarrow j+1] \\
C_{w_{k-1} x_{l-1}}^{u_{i-1} v_j} + AD(u_i, \frown, w_k, x_l) & [i \leftarrow i+1; j \leftarrow j+1; l \leftarrow l+1] \\
C_{w_k x_{l-1}}^{u_{i-1} v_j} + AD(u_i, \frown, \frown, x_l) & [i \leftarrow i+1; l \leftarrow l+1] \\
C_{w_{k-1} x_l}^{u_{i-1} v_j} + AD(u_i, \frown, w_k, \frown) & [i \leftarrow i+1; k \leftarrow k+1] \\
C_{w_k x_l}^{u_{i-1} v_j} + AD(u_i, \frown, \frown, \frown) & [i \leftarrow i+1] \\
C_{w_{k-1} x_{l-1}}^{u_i v_j} + AD(\frown, \frown, w_k, x_l) & [k \leftarrow k+1; l \leftarrow l+1] \\
C_{w_k x_{l-1}}^{u_i v_j} + AD(\frown, \frown, \frown, x_l) & [l \leftarrow l+1] \\
C_{w_{k-1} x_l}^{u_i v_j} + AD(\frown, \frown, w_k, \frown) & [k \leftarrow k+1]
\end{cases}
$$

**End**

When $i = |u|$ and $j = |v|$ and $k = |w|$ and $l = |x|$.

**Result**

$C_{w_{|w|} x_{|x|}}^{u_{|u|} v_{|v|}}$ is $AD(u, v, w, x)$ in $\Sigma^\star$.

**Complexity**

This algorithms runs in a time complexity in $\mathcal{O}\big(|u|.|v|.|w|.|x|\big)$.

**Correctness**

The correctness of this algorithm is demonstrated by recurrence, since it uses the dynamic programming principles. It requires only the analogical dissimilarity in $\Sigma'$ to have the properties that we have called: *coherence with analogy*, *symmetry for 'as'* and *exchange of medians*. The *triangle inequality* property is not necessary.

### 3.6 Generalized Resolution of Analogical Equations in Sequences: the SOLVANA Algorithm

#### 3.6.1 APPROXIMATE SOLUTIONS TO AN ANALOGICAL EQUATION

Up to now, we have considered that an analogical equation has either one (or several) exact solutions, or no solution. In the latter case, the concept of analogical dissimilarity is useful to define an approximate solution.

**Definition 3.5 (Best approximate solution to an analogical equation)** *Let $X$ be a set on which is defined an analogy and an analogical dissimilarity $AD$. Let $\quad a : b :: c : x \quad$ be an analogical equation in $X$. The set of best approximate solutions to this equation is given by:*

$$\left\{ y \ : \ \arg\min_{y \in X} AD(a, b, c, y) \right\}$$

In other words, the best approximate solutions are the objects $y \in X$ that are the closest to be in analogical proportion with $a$, $b$ and $c$. Obviously, this definition generalizes that of a solution to an analogical equation given at section 2.2. Since we have defined $AD$ with good properties on several alphabets and on sequences on these alphabets, we can compute an approximate solution to analogical equations in all these domains.

We can easily enlarge this concept and define the set of the $k$-best solutions to the analogical equation $\quad a : b :: c : x \quad$. Informally, it is any subset of $k$ elements of $X$ which have a minimal $AD$ when associated in fourth position with $a$, $b$ and $c$.

In $\mathbb{R}^n$ and $\{0, 1\}^n$, there is only one best approximate solution to an analogical equation, which can be easily computed (see sections 3.2 and 3.3). Finding the set of the $k$-best solutions is also a simple problem.

Let us turn now to an algorithm which finds the set of the best approximate solutions to the equation $\quad u : v :: w : x \quad$ when the objects are sequences on an alphabet on which an $AD$ has been defined. We will also make some comments to extend its capacity to find the set of the $k$-best solutions.

#### 3.6.2 THE SOLVANA ALGORITHM

This algorithm uses dynamic programming to construct a 3-dimensional array. When the construction is finished, a backtracking is performed to produce the DAG of all the best solutions.

An alignment of four sequences of different lengths is realized by inserting letters $\backsim$ so that all the four sequences have the same length. Once this is done, we consider in each column of the alignment the analogical dissimilarity in the augmented alphabet.

We construct a three dimensional $n_1 \times n_2 \times n_3$ matrix $M$ (respectively the length of the first, second and third sequences $A$, $B$ and $C$ of the analogical equation '$A$ is to $B$ as $C$ is to $x$'). To find the fourth sequence, we fill up $M$ with the following recurrence:

$$
\underset{1 \le i,j,k \le n_1,n_2,n_3}{M[i,j,k]} = Min \begin{cases} M[i-1,j-1,k-1] + \underset{x \in \Sigma'}{Min} \ AD(a_i,b_j,c_k,x) \\ M[i,j-1,k-1] + \underset{x \in \Sigma'}{Min} \ AD(\frown,b_j,c_k,x) \\ M[i,j,k-1] + \underset{x \in \Sigma'}{Min} \ AD(\frown,\frown,c_k,x) \\ M[i,j-1,k] + \underset{x \in \Sigma'}{Min} \ AD(\frown,b_j,\frown,x) \\ M[i-1,j,k-1] + \underset{x \in \Sigma'}{Min} \ AD(a_i,\frown,c_k,x) \\ M[i-1,j-1,k] + \underset{x \in \Sigma'}{Min} \ AD(a_i,b_j,\frown,x) \\ M[i-1,j,k] + \underset{x \in \Sigma'}{Min} \ AD(a_i,\frown,\frown,x) \end{cases}
$$

$a_i$ is the $i^{th}$ object of the sequence $A$. $\Sigma' = \Sigma \cup \{\frown\}$.

At each step, we save in the cell $M[i,j,k]$ not only the cost but also the letter(s) found by analogical resolution along the optimal way of progression. When $M$ is completed, a backward propagation gives us all the optimal generated sequences with the same optimal analogical dissimilarity, strucured as a DAG.

The computational complexity of this algorithm is $O(m * n^3)$, where $m = Card(\Sigma')$ and $n$ is the maximum length of sequences.

### 3.6.3 EXAMPLE

Let $\Sigma = \{a,b,c,A,B,C\}$ be an alphabet defined by 5 binary features, as follows:

|   | $f_1$ | $f_2$ | $f_3$ | $f_4$ | $f_5$ |
|---|---|---|---|---|---|
| $a$ | 1 | 0 | 0 | 1 | 0 |
| $b$ | 0 | 1 | 0 | 1 | 0 |
| $c$ | 0 | 0 | 1 | 1 | 0 |
| $A$ | 1 | 0 | 0 | 0 | 1 |
| $B$ | 0 | 1 | 0 | 0 | 1 |
| $C$ | 0 | 0 | 1 | 0 | 1 |
| $\frown$ | 0 | 0 | 0 | 0 | 0 |

The first three features indicates what is the letter (for example, $f_1$ is true on $a$ and $A$ only) and the last two indicate the case of the letter ($f_4$ holds true for lower case letters, $f_5$ for upper case letters).

For example, let $ab : Bc :: Bc : x$ be an analogical equation. There is no exact solution, but six best approximate solutions $y$ such that $AD(ab, Bc, Bc, y) = 4$, for example $y = BB$ or $y = Cc$. Figure 3 displays the DAG of the results produced by SOLVANA on this example.

## 4. Analogical Dissimilarity and Machine Learning

### 4.1 Motivation

We assume here that there exists an analogy defined on the set $X$ and an analogical dissimilarity $AD$ with the following properties: coherence with analogy, symmetry for 'as', triangle inequality, exchange of medians and asymmetry for 'is to'.

Let $\mathcal{S}$ be a set of elements of $X$, which is of cardinality $m$, and let $y$ be another element of $X$ with $y \notin \mathcal{S}$. The problem that we tackle in this section is to find the triple of objects $(u, v, w)$ in $\mathcal{S}$

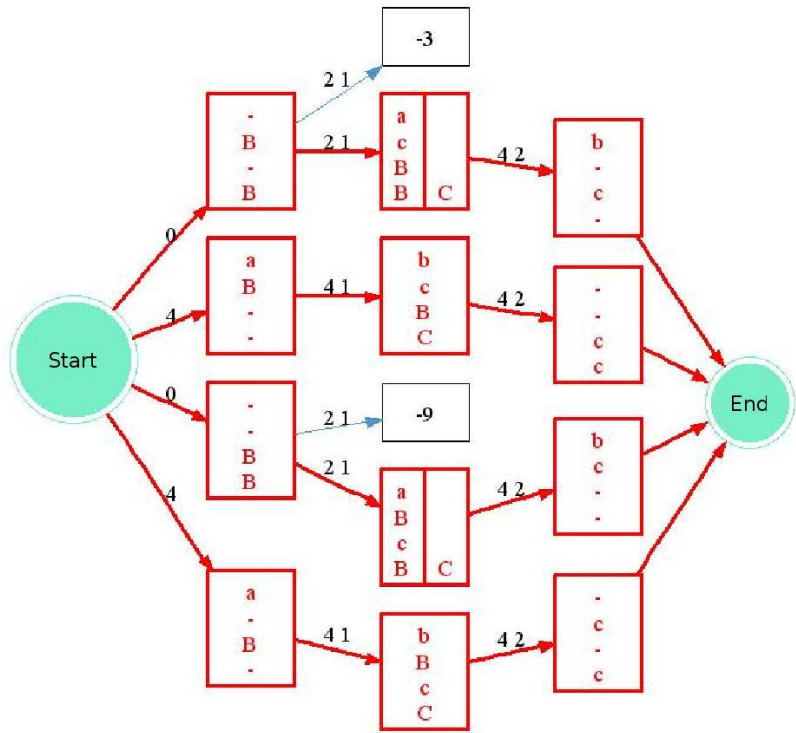

Figure 3: *Result of SOLVANA: the DAG of all best approximate solutions to an analogical equation on sequences. Each path displays a different alignment of optimal cost.*

such that:

$$AD(u, v, w, y) = \underset{t_1, t_2, t_3 \in \mathcal{S}}{\arg\min} \; AD(t_1, t_2, t_3, y)$$

This will directly lead us to use the notion of $AD$ in supervised machine learning, *e.g.* of a classification rule.

### 4.2 The Brute Force Solution

An obvious solution is to examine all the triples in $\mathcal{S}$. This brute force method requires $m^3$ calls to a procedure computing the analogical dissimilarity between four objects of $X$. According to the properties of analogical dissimilarity, this number can actually be divided by 8, but it does not change the theoretical and practical complexity of the search.

The situation is similar to that of the search for the nearest neighbor in Machine Learning, for which the naive algorithm requires $m$ distance computations. Many proposals have been made to decrease this complexity (see for example the work of Chávez, Navarro, Baeza-Yates, & Marroquín, 2001). We have chosen to focus on an extension of the AESA algorithm, based on the property of triangle inequality for distances (Micó, Oncina, & Vidal, 1994). Since we have defined the concept of analogical dissimilarity with a similar property, it is natural to explore how to extend this algorithm.

### 4.3 'FADANA': FAst search of the least Dissimilar ANAlogy

This section describes a fast algorithm to find, given a set of objects $\mathcal{S}$ of cardinality $m$ and an object $y$, the three objects $(z^\star, t^\star, x^\star)$ in $\mathcal{S}$ such that the analogical dissimilarity $AD(z^\star, t^\star, x^\star, y)$ is minimal. It is based on the AESA technique, which can be extended to analogical dissimilarity. Thanks to its properties, an analogical dissimilarity $AD(z, t, x, y)$ can be seen as a distance between the two couples $(z, t)$ and $(x, y)$, and consequently we will basically work on couples of objects. We use equivalently in this paragraph the terms '(analogical) distance between the two couples $(u, v)$ and $(w, x)$' and '(analogical) dissimilarity between the four elements $u, v, w$ and $x$' to describe $AD(u, v, w, x)$.

#### 4.3.1 PRELIMINARY COMPUTATION

In this part, which is done off line, we have to compute the analogical dissimilarity between every four objects in the data base. This step has a complexity in time and space of $\mathcal{O}(m^4)$, where $m$ is the size of $\mathcal{S}$. We will come back to this point in section 4.4, where we will progress from an $AESA$-like to a $LAESA$-like technique and reduce the computational complexity.

#### 4.3.2 PRINCIPLE OF THE ALGORITHM

The basic operation is to compose a couple of objects by adding to $y$ an object $x_i \in \mathcal{S}$ where $i = 1, m$. The goal is now to find the couple of objects in $\mathcal{S}$ having the lowest distance with $(x_i, y)$, then to change $x_i$ into $x_{i+1}$. Looping $m$ times on an *AESA*-like select and eliminate technique insures to finally find the triple in $\mathcal{S}$ having the lowest analogical dissimilarity when associated with $y$.

#### 4.3.3 NOTATIONS

Let us denote:

- $\mathcal{C}$ the set of couples $(u, v)$ which distance to $(x_i, y)$ has already been computed.

- $\delta = \underset{(z,t) \in \mathcal{U}}{\arg\min}(AD(z, t, x_i, y))$

- $\delta_i = \underset{(z,t) \in \mathcal{U}, 1 \le j \le i}{\arg\min}(AD(z, t, x_i, y))$

- $Dist = \{AD(z, t, x_i, y), (z, t) \in \mathcal{C}\}$

- $Dist(j)$ the $j^{th}$ element of $Dist$

- $Quad_{\mathcal{U}} = \{(z, t, x_i, y), (z, t) \in \mathcal{C}\}$

- $Quad_{\mathcal{U}}(j)$ the $j^{th}$ element of $Quad_{\mathcal{U}}$

The algorithm is constructed in the three following phases:

#### 4.3.4 INITIALIZATION

Each time that $x_i$ changes (when $i$ is increased by 1), the set $\mathcal{U}$ is refilled with all the possible couples of objects $\in \mathcal{S}$.

The set $\mathcal{C}$ and $Dist$ which contain respectively the couples and the distances to $(x_i, y)$ that have been measured during one loop, are initialized as empty sets.

The local minimum $Min$, containing the minimum of analogical dissimilarities of one loop is set to infinity.

$k = Card(\mathcal{C})$ represents the number of couples where the distance have been computed with $(x_i, y)$ in the current loop. $k$ is initialized to zero.

---

**Algorithm 1** Algorithm FADANA: initialization.

---

**begin**
$\mathcal{U} \leftarrow \{(x_i, x_j),\ i = 1, m \text{ and } j = 1, m\}$;
$\mathcal{C} \leftarrow \varnothing$;
$Min \leftarrow +\infty$;
$Dist \leftarrow \varnothing$;
$k \leftarrow 0$;
**end**

---

### 4.3.5 SELECTION

The goal of this function is to extract from the set $\mathcal{U}$ the couple $(zz, tt)$ that is the more promising in terms of the minimum analogical dissimilarity with $(x_i, y)$, using the criterion:

$$(zz, tt) = \arg\min_{(u,v) \in \mathcal{U}} \ \underset{(z,t) \in C}{Max} \ \Big|\ AD(u, v, z, t) - AD(z, t, x_i, y)\ \Big|$$

---

**Algorithm 2** Algorithm FADANA: selection of the most promising couple.

---

$\texttt{selection}(\mathcal{U}, \mathcal{C}, (x_i, y), Dist)$
**begin**
$s \leftarrow 0$
**for** $i = 1, Card(\mathcal{U})$ **do**
    **if** $s \leq \sum_{j \in \mathcal{C}} |AD(z_j,\ t_j,\ u_i,\ v_i) - Dist(j)|$ **then**
        $s \leftarrow \sum_{j \in \mathcal{C}} |AD(z_j,\ t_j,\ u_i,\ v_i) - Dist(j)|$;
        $\arg\min \leftarrow i$;
    **end if**
**end for**
Return $(u_{\arg\min},\ v_{\arg\min})$;
**end**

---

### 4.3.6 ELIMINATION

During this section all the couples $(u, v) \in \mathcal{U}$ where the analogical distance with $(x_i, y)$ can not be less than what we already found are eliminated thanks to the two criteria below:

$$AD(u, v, z, t) \leq AD(z, t, y, x_i) - \delta \Rightarrow AD(u, v, x_i, y) \geq \delta$$

and

$$AD(u, v, z, t) \geq AD(z, t, y, x_i) + \delta \Rightarrow AD(u, v, x_i, y) \geq \delta$$

where $\delta = AD(z^\star, t^\star, x^\star, y)$ represents the minimum analogical dissimilarity found until now (see figure 4). Note that $\delta$ is updated during the whole algorithm and is never reinitialized when $i$ is increased.

---

**Algorithm 3** Algorithm FADANA: elimination of the useless couples.

---

`eliminate`$(\mathcal{U}, \mathcal{C}, (x_i, y), \delta, k)$
$(z_k,\ t_k)$ is the $k^{th}$ element of $Quad_\mathcal{U}$
**begin**
**for** $i = 1, Card(\mathcal{U})$ **do**
    **if** $AD(z_k,\ t_k,\ u_i,\ v_i) \leq Dist(k) + \delta$ **then**
        $\mathcal{U} \leftarrow \mathcal{U} - \{(u_i,\ v_i)\};$
        $\mathcal{C} \leftarrow \mathcal{C} \cup \{(u_i,\ v_i)\};$
    **else if** $AD(z_k,\ t_k,\ u_i,\ v_i) \geq Dist(k) - \delta$ **then**
        $\mathcal{U} \leftarrow \mathcal{U} - \{(u_i,\ v_i)\};$
        $\mathcal{C} \leftarrow \mathcal{C} \cup \{(u_i,\ v_i)\};$
    **end if**
**end for**
**end**

---

**Algorithm 4** Algorithm FADANA: main procedure.

---

**begin**
$\mathcal{S} \leftarrow \{x_i,\ i = 1, m\};$
$AD^\star \leftarrow +\infty;$
**for** $i = Card(\mathcal{S})$ **do**
   `Initialize`;
   **while** $\mathcal{U} \neq \varnothing$ **do**
      $(z, t) \leftarrow$ `selection`$(\mathcal{U}, \mathcal{C}, (x_i, y), Dist);$
      $Dist(k) \leftarrow AD(z, t, x_i, y);$
      $k = k + 1;$
      $\mathcal{U} \leftarrow \mathcal{U} - \{(z, t)\};$
      $\mathcal{C} \leftarrow \mathcal{C} \cup \{(z, t)\};$
      **if** $Dist(k) \geq Min$ **then**
         `eliminate`$(\mathcal{U}, \mathcal{C}, (x_i, y), \delta, k)$
      **else**
         $Min \leftarrow Dist(k);$
         **if** $Dist(k) < AD^\star$ **then**
             $AD^\star \leftarrow Dist(k);$
             $z^\star \leftarrow z,\ t^\star \leftarrow t,\ x^\star \leftarrow x_i;$
         **end if**
         **for** $k = 1, Card(\mathcal{C})$ **do**
            `eliminate`$(\mathcal{U}, \mathcal{C}, (x_i, y), \delta, k)$
         **end for**
      **end if**
   **end while**
**end for**
The best triple in $\mathcal{S}$ is $(z^\star, t^\star, x^\star)$ ;
The least analogical dissimilarity is $AD^\star = AD(z^\star, t^\star, x^\star, y)$ ;
**end**

---

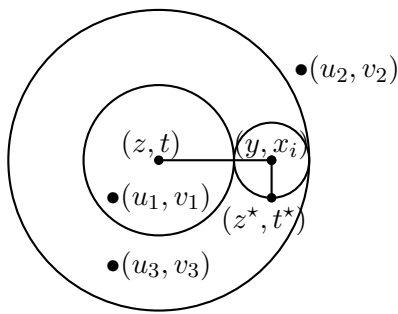

Figure 4: Elimination process in FADANA.

### 4.4 Selection of Base Prototypes in FADANA

So far, *FADANA* has the drawback of requiring a precomputing time and storage in $\mathcal{O}(m^4)$, which is in practice impossible to handle for $m > 100$.

To go further, we have devised an ameliorated version of the FADANA algorithm, in which the preliminary computation and storage is limited to $N.m^2$, where $N$ is a certain number of *couples* of objects. The principle is similar to that of $LAESA$ (Micó et al., 1994). $N$ *base prototypes* couples are selected among the $m^2$ possibilities through a greedy process, the first one being chosen at random, the second one being as far as possible from the first one, and so on. The distance between couples of objects is, according to the definition of the analogical dissimilarity:

$$\delta\big((x,y),(z,t)\big) = AD(z,t,x,y)$$

### 4.5 Efficiency of FADANA

We have conducted some experiments to measure the efficiency of FADANA. We have tested this algorithm on four databases from the UCI Repository (Newman, Hettich, Blake, & Merz, 1998), by noting each time the percentage of $AD$ computed in-line for different numbers of base prototypes compared to those made by the naive method (see Figure 5, the scales are logarithmic). The number of base prototypes is expressed as percentage on the learning set. Obviously, if the learning set contains $m$ elements, the number of possible 3-tuples that can be built is $m^3$. This point explains why the percentage of base prototypes compared to the size of the learning set can rise above 100%. The number of in-line computations of the $AD$ is the mean over the test set.

We observe in these results that the optimal number of base prototypes is between 10% and 20% if we aim to optimize the computation time performance.

## 5. Two Applications in Machine Learning Problems

### 5.1 Classification of Objects Described by Binary and Nominal Features

The purpose of this first experiment is to measure the benefit of analogical dissimilarity applied to a basic problem of classification, compared to standard classifiers such $k$-nearest neighbors, neural networks, and decision trees. In this benchmarking, we are not yet interested in classifying sequences, but merely to investigate what the basic concept of analogical dissimilarity can bring to the learning of a classification rule for symbolic objects.

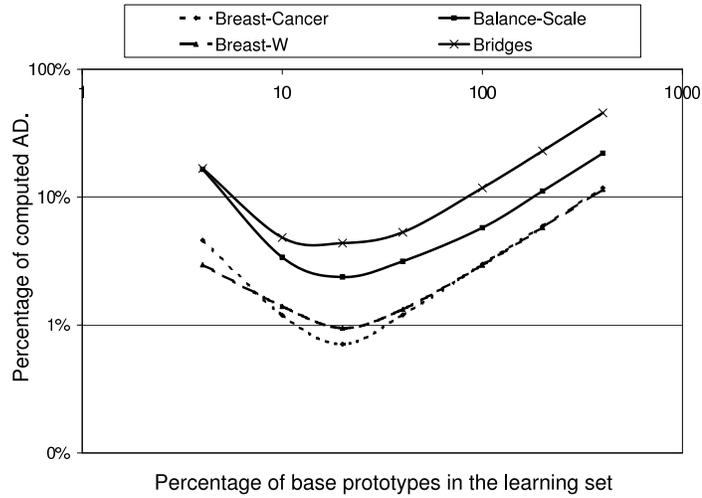

Figure 5: Efficiency of FADANA w.r.t. the number of base prototypes

### 5.1.1 METHOD DESCRIPTION

Let $\mathcal{S} = \big\{ \big(o_i, h(o_i)\big) \mid 1 \le i \le m \big\}$ be a learning set, where $h(o_i)$ is the class of the object $o_i$. The objects are defined by binary attributes. Let $x$ be an object not in $\mathcal{S}$. The learning problem is to find the class of a new object $x$, using the learning set $\mathcal{S}$. To do this, we define a learning rule based on the concept of analogical dissimilarity depending on an integer $k$, which could be called the $k$ *least dissimilar 3-tuple* rule.

The basic principle is the following: among all the 3-tuples $(a, b, c)$ in $\mathcal{S}^3$, we consider the subset of those which produce the least analogical dissimilarity when associated with $x$ (the FADANA algorithm is used here). For a part of them, the analogical equation $h(a) : h(b) :: h(c) : g$ has an exact solution in the finite set of the classes. We keep only these 3-tuples and we choose the class which takes the majority among these values $g$ as the class for $x$.

More precisely, the procedure is as follows:

1. Compute the analogical dissimilarity between $x$ and all the $n$ 3-tuples in $\mathcal{S}$ which produce a solution for the class of $x$.

2. Sort these $n$ 3-tuples by the increasing value of their $AD$ when associated with $x$.

3. If the $k$-th object has the value $p$, then let $k'$ be the greatest integer such that the $k'$-th object has the same value $p$.

4. Solve the $k'$ analogical equations on the label of the class. Take the winner of the votes among the $k'$ results.

To explain, we firstly consider the case where there are only two classes $\omega_0$ and $\omega_1$. An example with 3 classes will follow.

Point 1 means that we retain only the 3-tuples which have one of the four[4] configurations for their class displayed in Table 1. We ignore the 3-tuples that do not lead to an equation with a trivial solution on classes:

---

4. There are actually two more, each one equivalent to one of the four (by exchange of the means objects).

| $h(a)$ | : | $h(b)$ | :: | $h(c)$ | : | $h(x)$ | : | ? | resolution |
|--------|---|--------|----|--------|---|--------|---|---|------------|
| $\omega_0$ | : | $\omega_0$ | :: | $\omega_0$ | : | ? | | | $h(x) = \omega_0$ |
| $\omega_1$ | : | $\omega_0$ | :: | $\omega_1$ | : | ? | | | |
| $\omega_1$ | : | $\omega_1$ | :: | $\omega_1$ | : | ? | | | $h(x) = \omega_1$ |
| $\omega_0$ | : | $\omega_1$ | :: | $\omega_0$ | : | ? | | | |

Table 1: Possible configurations of a 3-tuple

| $o_1$ | $o_2$ | $o_3$ | $h(o_1)$ | $h(o_2)$ | $h(o_3)$ | $h(x)$ | $AD$ | $k$ |
|-------|-------|-------|----------|----------|----------|--------|------|-----|
| $b$ | $a$ | $d$ | $\omega_0$ | $\omega_0$ | $\omega_1$ | $\omega_1$ | 0 | 1 |
| $b$ | $d$ | $e$ | $\omega_0$ | $\omega_1$ | $\omega_2$ | $\bot$ | 1 | |
| $c$ | $d$ | $e$ | $\omega_1$ | $\omega_1$ | $\omega_2$ | $\omega_2$ | 1 | 2 |
| $a$ | $b$ | $d$ | $\omega_0$ | $\omega_0$ | $\omega_1$ | $\omega_1$ | 1 | 3 |
| $c$ | $a$ | $e$ | $\omega_1$ | $\omega_0$ | $\omega_2$ | $\bot$ | 2 | |
| $d$ | $c$ | $e$ | $\omega_1$ | $\omega_1$ | $\omega_2$ | $\omega_2$ | 2 | 4 |
| $d$ | $b$ | $c$ | $\omega_1$ | $\omega_0$ | $\omega_1$ | $\omega_0$ | 2 | 5 |
| $a$ | $c$ | $e$ | $\omega_0$ | $\omega_1$ | $\omega_2$ | $\bot$ | 2 | |
| $a$ | $c$ | $c$ | $\omega_0$ | $\omega_1$ | $\omega_1$ | $\bot$ | 3 | |
| $a$ | $b$ | $e$ | $\omega_0$ | $\omega_0$ | $\omega_2$ | $\omega_2$ | 3 | 6 |
| $b$ | $a$ | $e$ | $\omega_0$ | $\omega_0$ | $\omega_2$ | $\omega_2$ | 3 | 7 |
| $b$ | $c$ | $d$ | $\omega_0$ | $\omega_1$ | $\omega_1$ | $\bot$ | 3 | |
| $c$ | $c$ | $c$ | $\omega_1$ | $\omega_1$ | $\omega_1$ | $\omega_1$ | 4 | 8 |
| $a$ | $a$ | $c$ | $\omega_0$ | $\omega_0$ | $\omega_1$ | $\omega_1$ | 4 | 9 |
| ... | ... | ... | ... | ... | ... | ... | ... | |

Table 2: An example of classification by analogical dissimilarity. Analogical proportions whose analogical resolution on classes have no solution (represented by $\bot$) are not taken into account. $AD$ is short for $AD(o_1, o_2, o_3, x)$.

| $h(a)$ | : | $h(b)$ | :: | $h(c)$ | : | $h(x)$ |
|--------|---|--------|----|--------|---|--------|
| $\omega_0$ | : | $\omega_1$ | :: | $\omega_1$ | : | ? |
| $\omega_1$ | : | $\omega_0$ | :: | $\omega_0$ | : | ? |

**Example**

Let $\mathcal{S} = \{(a, \omega_0), (b, \omega_0), (c, \omega_1), (d, \omega_1), (e, \omega_2)\}$ be a set of five labelled objects and let $x \notin \mathcal{S}$ be some object to be classified. According to the analogical proportion axioms, there is only 75 ($= (Card(\mathcal{S})^3 + Card(\mathcal{S})^2)/2$) non-equivalent analogical equations among $125 (= Card(\mathcal{S})^3)$ equations that can be formed between three objects from $\mathcal{S}$ and $x$. Table (2) shows only the first 14 lines after sorting with regard to some arbitrarily analogical dissimilarity. The following table gives the classification of an object $x$ according to $k$:

| $k$ | 1 | 2 | 3 | 4 | 5 | 6 | 7 |
|-----|---|---|---|---|---|---|---|
| $k'$ | 1 | 3 | 3 | 5 | 5 | 7 | 7 |
| classification of $x$ | 1 | 1 | 1 | ? | ? | 2 | 2 |

### 5.1.2 WEIGHTING THE ATTRIBUTES

The basic idea in weighting the attributes is that they do not have the same importance in the classification, and that more importance has to be given to the most discriminative. The idea of selecting or enhancing interesting attributes is classical in Machine Learning, and not quite new in the framework of analogy. In a paper of Turney (2005), a discrimination is done by keeping the most frequent patterns in words. Therefore, a greater importance is given to the attributes that are actually discriminant. However, in an analogical classification system, there are several ways to find the class of the unknown element. Let us take again the preceding two class problem example (see table 1) to focus on this point.

We notice that there are two ways to decide between the class $\omega_0$ and the class $\omega_1$ (there is also a third possible configuration which is equivalent to the second by exchange of the means). We therefore have to take into account the equation used to find the class. This is why we define a set of weights for each attribute, depending on the number of classes. These sets are stored in what we call an *analogical weighting matrix*.

**Definition 5.1** *An analogical weighting matrix (W) is a three dimensional array. The first dimension is for the attributes, the second one is for the class of the first element in an analogical proportion and the third one is for the class of the last element in an analogical proportion. The analogical proportion weighting matrix is a $d \times C \times C$ matrix, where $d$ is the number of attributes and $C$ is the number of classes.*

*For a given attribute $a_k$ of rank $k$, the element $W_{kij}$ of the matrix indicates which weight must be given to $a_k$ when encountered in an analogical proportion on classes whose first element is $\omega_i$, and for which $\omega_j$ is computed as the solution.*

Hence, for the attribute $a_k$:

|  |  | Last element (decision) | |
|---|---|---|---|
|  |  | class $\omega_i$ | class $\omega_j$ |
| First ele- | class $\omega_i$ | $W_{kii}$ | $W_{kij}$ |
| ment | class $\omega_j$ | $W_{kji}$ | $W_{kjj}$ |

Since we only take into account the 3-tuples that give a solution on the class decision, all the possible situations are of one of the three patterns:

| Possible patterns | First element | Decision class |
|---|---|---|
| $\omega_i \;:\; \omega_i \;::\; \omega_j \;:\; \omega_j$ | $\omega_i$ | $\omega_j$ |
| $\omega_i \;:\; \omega_j \;::\; \omega_i \;:\; \omega_j$ | $\omega_i$ | $\omega_j$ |
| $\omega_i \;:\; \omega_i \;::\; \omega_i \;:\; \omega_i$ | $\omega_i$ | $\omega_i$ |

This observation gives us a way to compute the values $W_{kij}$ from the learning set.

### 5.1.3 LEARNING THE WEIGHTING MATRIX FROM THE TRAINING SAMPLE

The goal is now to fill the three dimensional analogical weighting matrix using the learning set. We estimate $W_{kij}$ by the frequency that the attribute $k$ is in an analogical proportion with the first element class $\omega_i$, and solves in class $\omega_j$.

Firstly, we tabulate the splitting of every attribute $a_k$ on the classes $\omega_i$:

|         | ... class $\omega_i$ ... |          |
|---------|:------------------------:|----------|
| $a_k = 0$ | ...                    | $n_{0i}$ | ... |
| $a_k = 1$ | ...                    | $n_{1i}$ | ... |

where $a_k$ is the attribute $k$ and $n_{0i}$ (resp. $n_{1i}$) is the number of objects in the class $i$ that have the value 0 (resp. 1) for the binary attribute $k$. Hence, $\sum_{k=0}^{1} \sum_{i=1}^{C} n_{ki} = m$ (the number of objects in the training set). Secondly, we compute $W_{kij}$ by estimating the probability to find a correct analogical proportion on attribute $k$ with first element class $\omega_i$ which solves in class $\omega_j$.

In the following table we show all the possible ways of having an analogical proportion on the binary attribute $k$. $0_i$ (resp. $1_i$) is the 0 (resp. 1) value of the attribute $k$ that has class $\omega_i$.

| $1^{st}$ | $0_i : 0_i :: 0_j : 0_j$ | $4^{th}$ | $1_i : 1_i :: 1_j : 1_j$ |
|----------|--------------------------|----------|--------------------------|
| $2^{sd}$ | $0_i : 1_i :: 0_j : 1_j$ | $5^{th}$ | $1_i : 0_i :: 1_j : 0_j$ |
| $3^{rd}$ | $0_i : 0_i :: 1_j : 1_j$ | $6^{th}$ | $1_i : 1_i :: 0_j : 0_j$ |

$\mathcal{P}_k(1^{st})$ estimates the probability that the first analogical proportion in the table above occurs.

$$\mathcal{P}_k(1^{st}) = n_{0i}n_{0i}n_{0j}n_{0j}/m^4$$
$$\vdots$$

From $W_{kij} = \mathcal{P}_k(1^{st}) + \cdots + \mathcal{P}_k(6^{th})$, we compute

$$W_{kij} = \left((n_{0i}^2 + n_{1i}^2)(n_{0j}^2 + n_{1j}^2) + 2 * n_{0i}n_{0j}n_{1i}n_{1j}\right)/(6 * m^4)$$

The decision algorithm of section 5.1.1 is only modified at point 1, which turns into *Weighted Analogical Proportion Classifier* ($WAPC$):

- Given $x$, find all the $n$ 3-tuples in $\mathcal{S}$ which can produce a solution for the class of $x$. For every 3-tuple among these $n$, say $(a, b, c)$, consider the class $\omega_i$ of the first element $a$ and the class $\omega_j$ of the solution. Compute the analogical dissimilarity between $x$ and this 3-tuple with the weighted $AD$:

$$AD(a, b, c, x) = \sum_{k=1}^{d} W_{kij} AD(a_k, b_k, c_k, x_k)$$

Otherwise, if point 1 is not modified, the method is called *Analogical Proportion Classifier* ($APC$).

### 5.1.4 EXPERIMENTS AND RESULTS

We have applied the weighted analogical proportion classifier ($WAPC$) to eight classical data bases, with binary and nominal attributes, of the UCI Repository.

**MONK 1,2** and **3** Problems (*MO.1*, *MO.2* and *MO.3*), MONK3 problem has noise added. **SPECT** heart data (*SP.*). **Balance-Scale** (*B.S*) and **Hayes Roth** (*H.R*) database, both multiclass database. **Breast-W** (*Br.*) and **Mushroom** (*Mu.*), both data sets contain missing values. **kr-vs-kp** Kasparov vs Karpov (*k.k.*).

In order to measure the efficiency of $WAPC$, we have applied some standard classifiers to the same databases, and we have also applied $APC$ to point out the contribution of the weighting matrix (Sect.5.1.2). We give here the parameters used for the comparison method in Table 3:

- **Decision Table**: the number of non improving decision tables to consider before abandoning the search is 5.

- **Id3**: unpruned decision tree, no missing values allowed.

- **Part**: partial C4.5 decision tree in each iteration and turns the 'best' leaf into a rule, One-per-value encoding.

- **Multi layer Perceptron**: back propagation training, One-per-value encoding, one hidden layer with (# classes + # attributes)/2 nodes.

- **LMT** ('logistic model trees'): classification trees with logistic regression functions at the leaves, One-per-value encoding.

- **IB1**: Nearest-neighbor classifier with normalized Euclidean distance, which have better results than **IB10**.

- **JRip**: propositional rule learner, Repeated Incremental Pruning to Produce Error Reduction (RIPPER), optimized version of IREP. .

We have worked with the WEKA package (Witten & Frank, 2005), choosing 6 different classification rules on the same data. Some are well fit to binary data, like ID3, PART, Decision Table. Others, like IB1 or Multilayer Perceptron, are more adapted to numerical and noisy data.

The results are given in Table 3. We have arbitrarily taken $k = 100$ for our two rules. The value $k$ is not very sensitive in the case of nominal and binary data and on small databases such as the ones that are used in the experiments (see Figure 6). However, it is possible to set $k$ using a validation set.

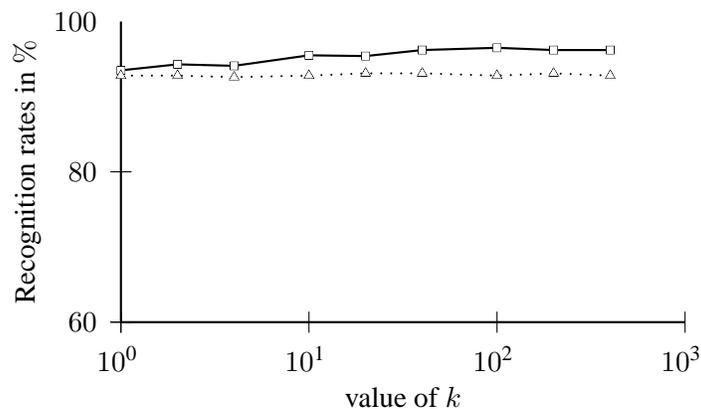

Figure 6:  Modification in recognition rate subject to $k$. Full line and dotted line are respectively the recognition rates on the database 'breast-w' and 'vote'.

We draw the following conclusions from this study: firstly, according to the good classification rate of $WAPC$ in *Br.* and *Mu.* databases, we can say that $WAPC$ handles the missing values well. Secondly, $WAPC$ seems to belong to the best classifiers for the *B.S* and *H.R* databases, which confirms that $WAPC$ deals well with multiclass problems. Thirdly, as shown by the good classification rate of $WAPC$ in the *MO.3* problem, $WAPC$ handles well noisy data. Finally, the results on *MO.* and *B.S* database are exactly the same with the weighted decision rule $WAPC$ than with $APC$. This is due to the fact that all $AD$ that are computed up to $k = 100$ are of null value. But on the other data bases, the weighting is quite effective. Unfortunaly, the last database show that $WAPC$ have a poor recognition rate on some databases, which means that analogy do not fit all classification problems.

| Methods | $MO.1$ | $MO.2$ | $MO.3$ | $SP.$ | $B.S$ | $Br.$ | $H.R$ | $Mu.$ | $k.k.$ |
|---|---|---|---|---|---|---|---|---|---|
| nb. of nominal atts. | 7 | 7 | 7 | 22 | 4 | 9 | 4 | 22 | 34 |
| nb. of binary atts. | 15 | 15 | 15 | 22 | 4 | 9 | 4 | 22 | 38 |
| nb. of train instances | 124 | 169 | 122 | 80 | 187 | 35 | 66 | 81 | 32 |
| nb. of test instances | 432 | 432 | 432 | 172 | 438 | 664 | 66 | 8043 | 3164 |
| nb. of classes | 2 | 2 | 2 | 2 | 3 | 2 | 4 | 2 | 2 |
| **WAPC** ($k = 100$) | **98%** | **100%** | **96%** | 79% | **86%** | **96%** | **82%** | **98%** | 61% |
| APC ($k = 100$) | **98%** | **100%** | **96%** | 58% | **86%** | **91%** | 74% | **97%** | 61% |
| Decision Table | **100%** | 64% | **97%** | 65% | 67% | 86% | 42% | **99%** | 72% |
| Id3 | 78% | 65% | **94%** | 71% | 54% | – | 71% | – | 71% |
| PART | 93% | 78% | **98%** | **81%** | 76% | 88% | **82%** | **94%** | 61% |
| Multi layer Perceptron | **100%** | **100%** | **94%** | 73% | **89%** | **96%** | 77% | **96%** | 76% |
| LMT | 94% | 76% | **97%** | **77%** | **89%** | 88% | **83%** | **94%** | 81% |
| IB1 | 79% | 74% | 83% | **80%** | 62% | **96%** | 56% | **98%** | 71% |
| IBk ($k = 10$) | 81% | 79% | 93% | 57% | 82% | 86% | 61% | 91% | – |
| IB1 ($k = 5$) | 73% | 59% | **97%** | 65% | 78% | **95%** | **80%** | **97%** | – |
| JRip | 75% | 62.5% | 88% | **80%** | 69% | 86% | **85%** | **97%** | **94%** |

Table 3: Comparison Table between $WAPC$ and other classical classifiers on eight data sets. Best classifiers on a database are in bold with a significance level equal to 5%.

## 5.2 Handwritten Character Recognition: Generation of New Examples

### 5.2.1 INTRODUCTION

In a number of Pattern Recognition systems, the acquisition of labeled data is expensive or user unfriendly process. For example, when buying a smartphone equipped with a handwritten recognition system, the customer is not likely to write dozens of examples of every letter and digit in order to provide the system with a consequent learning sample. However, to be efficient, any statistical classification system has to be retrained to the new personal writing style or the new patterns with as many examples as possible, or at least a sufficient number of well chosen examples.

To overcome this paradox, and hence to make possible the learning of a classifier with very few examples, a straightforward idea is to generate new examples by randomly adding noise to the elements of a small learning sample. In his recent book, Bishop (2007) gives no theoretical coverage of such a procedure, but rather draws a pragmatic conclusion: ' ... the addition of random noise to the inputs ... has been shown to improve generalization in appropriate circumstances'.

As far as character recognition is concerned, generating synthetic data for the learning of a recognition system has mainly be used with offline systems (which process an image of the char-

$$
\begin{aligned}
u &= \quad 9 \ \curvearrowright 9 \ \curvearrowright 9\,9\,9\,9\,9 \ \curvearrowright E \ 1 \ 2 \ \curvearrowright 4 \ L \ 6 \ 9 \ 9 \ 9 \\
v &= \quad 1 \ L \ \curvearrowright 8 \ 9\,9\,9\,9\,9 \ 10 \ E \ \curvearrowright 2 \ 2 \ 4 \ L \ \curvearrowright 8 \ 8 \ 9 \\
w &= \quad \curvearrowright \curvearrowright 9 \ 8 \ 9\,9\,9\,9\,9 \ 10 \ E \ 2 \ 2 \ 3 \ 3 \ L \ 8 \ 9 \ 9 \ \curvearrowright \\
x_0 &= \quad 1 \ L \ \curvearrowright 8 \ 9\,9\,9\,9\,9 \ 10 \ E \ 2 \ 2 \ 3 \ 3 \ L \ 8 \ 8 \ 8 \ \curvearrowright
\end{aligned}
$$

(a)

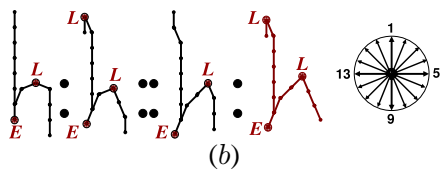

(b)

Figure 7: (a) Resolution on Freeman direction sequences by AP. (b) the corresponding characters representation.

acter). For offline character recognition, several image distortions have however been used (Cano, Pérez-Cortes, Arlandis, & Llobet, 2002): slanting, shrinking, ink erosion and ink dilatation. For online character recognition, several online distortions have been used, such as speed variation and angular variation (Mouchère, Anquetil, & Ragot, 2007).

We therefore are interested in the quick tuning of a handwritten character recognition to a new user, and we consider that only a very small set of examples of each character (typically 2 or 3) can be required from the new user. As we learn a writer-dependent system, the synthetic data have to keep the same handwriting style as the original data.

### 5.2.2 ANALOGY BASED GENERATION

In this second experiment, we are interested in handwritten characters, which are captured online. They are represented by a sequence of letters of $\Sigma$, where $\Sigma = \{1, 2, ..., 16, 0, C, ..., N\}$ is the alphabet of the Freeman symbols code augmented of symbols for *anchorage points*. These anchorage points come from an analysis of the stable handwriting properties, as defined in (Mouchère et al., 2007): pen-up/down, y-extrema, angular points and in-loop y-extrema.

Having a learning set that contains a few examples of each letter, we generate synthetic examples by analogical proportion as described in section 3.6 (see Figure 7). Hence, by generating artificial examples of the letter $f$ by analogical proportion using only three instances we augment the learning set with new and different examples as shown in the following pictures.

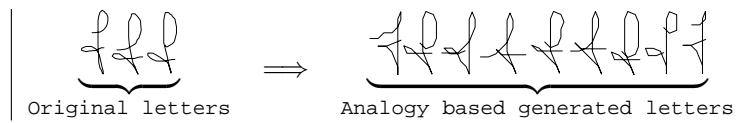

### 5.2.3 EXPERIMENTS

In this section we show that our generation strategies improves the recognition rate of three classical classifiers learned with few data.

**Experimental Protocol** In the data base that we use (Mouchère et al., 2007), twelve different writers have written 40 times the 26 lowercase letters (1040 characters) on a PDA. We use a 4-fold stratified cross validation. The experiments are composed of two phases in which three writer-dependent recognition systems are learned: a Radial Basis Function Network (RBFN), a K-Nearest Neighbor (K-NN) and a one-against-all Support Vector Machine (SVM).

Firstly, we compute two Reference recognition Rates without data generation: $RR10$ which is the recognition rate achievable with 10 original characters without character generation and $RR30$ gives an idea of achievable recognition rates with more original data. Practically speaking, in the context of on the fly learning phase we should not ask the user to input more than 10 characters per class.

Secondly the artificial character generation strategies are tested. For a given writer, one to ten characters per class are randomly chosen. Then 300 synthetic characters per class are generated to make a synthetic learning database. This experiment is done 3 times per cross validation split and per writer (12 times per user). The mean and the standard deviation of these 12 performance rates are computed. Finally the means of these measurements are computed to give a writer dependent mean recognition rate and the associated standard deviation.

We study three different strategies for the generation of synthetic learning databases. The strategy *'Image Distortions'* chooses randomly for each generation one among several image distortions. In the same way the strategy *'Online and Image Distortions'* chooses randomly one distortion among the image distortions and online distortions. The *'Analogy and Distortions'* strategy generates two-thirds of the base with the previous strategy and the remaining third with AP generation.

**Results**   Figure 8 compares the recognition rates achieved by the three generation strategies for the three classifiers. Firstly we can note that the global behavior is the same for the three classifiers. Thus the following conclusions do not depend on the classifier type. Secondly the three generation strategies are complementary because using *'Online and Image Distortions'* is better than *'Image Distortions'* alone and *'Analogy and Distortions'* is better than using distortions. Furthermore using only four original character with the complete generation strategy is better than the $RR10$. The $RR30$ is achieved by using 9 or 10 original characters. Thus we can conclude that using our generation strategies learns classifier with very few original data as efficiently as using original data from a long input phase : we need about three times fewer original data to achieve the same recognition rate.

Comparing 'Image Distortions', 'Online Distortions' and 'Analogy' alone shows that 'Analogy' is less efficient than the *ad-hoc* methods. Nevertheless, generating sequences by approximate analogical proportion is meaningful and somewhat independant of classical distorsions. In other words, the analogy of character shapes, which is used in 'natural intelligence', has been somehow captured by our definition and algorithms.

Our aim here is to know if the difference of the average of the three methods is significant. We have performed two methods of validation to evaluate the difference between two stategies. The first method is parametric: the T-TEST (Gillick & Cox, 1989). The second method is non-parametric: the SIGN TEST (Hull, 1993). In both methods, the comparaison is between the first and the second strategy then between the second and the third strategy on each number of original characters.

The T-TEST compares the value of the difference between the two generation methods regarding to the variation between the differences. The assumption is that the errors are in a normal distribution and that the errors are independent. If the mean difference is large comparing to the standard deviation, the two strategies are statistically different. In our case, the probability that our results are a random artefact is less than $10^{-12}$.

The Sign Test is non-parametric comparison method. Its benefit is to avoid assumptions on the normal distribution of the observations and the errors. This test replaces each difference by the sign of this difference. The sum of these occurrences is compared to the value of the hypothesis $H_0$ ($H_0$: the difference between the methods is not significant). Thus if a strategy is frequently better than the expected mean, then this strategy is significantly better. In our case, the probability that the hypothesis $H_0$ is true is less than $10^{-30}$. Hence, the difference is significantly better.

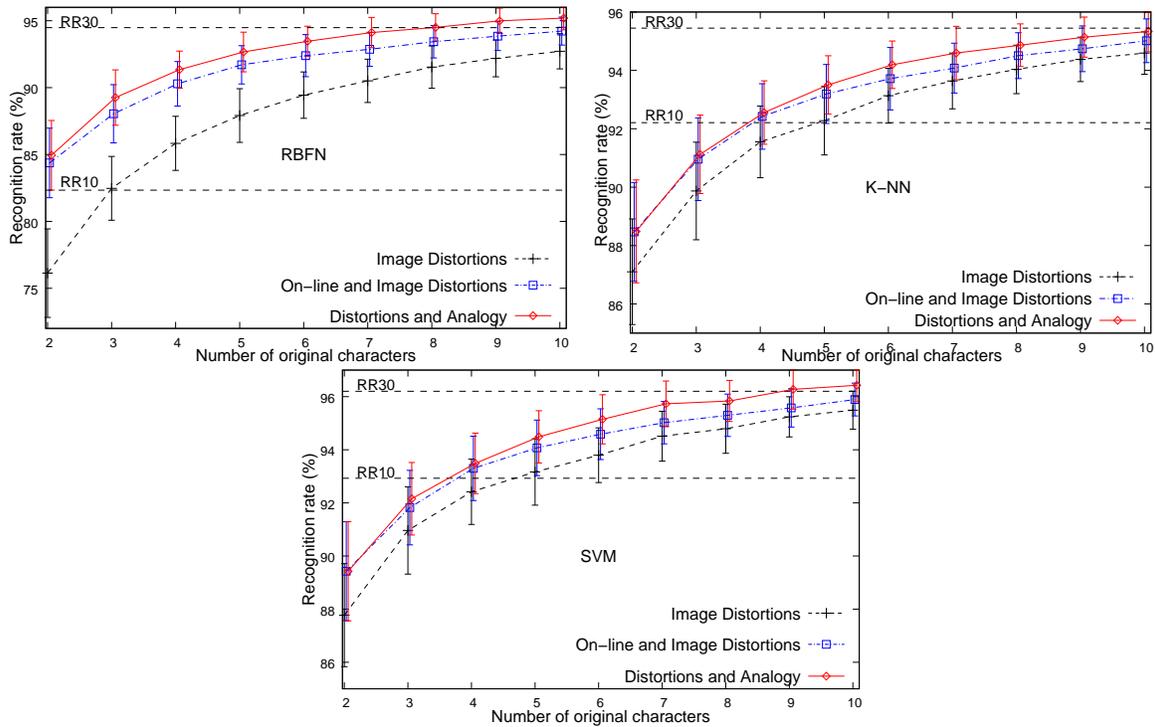

Figure 8: Writer-dependent recognition rates (mean and standard deviation) depending on the number of used original characters compared to reference rates using 10 or 30 characters per class for RBFN, KNN and SVM classifiers.

## 6. Conclusions and Future Work

In this article, we have investigated a formal notion of analogy between four objects in the same universe. We have given definitions of analogy, formulas and algorithms for solving analogical equations in some particular sets. We have given a special focus on objects structured as sequences, with an original definition of analogy based on optimal alignments. We also have introduced, in a coherent manner, the new notion of analogical dissimilarity, which quantifies how far four objects are from being in analogy. This notion is useful for lazy supervised learning: we have shown how the time consuming brute force algorithm could be ameliorated by generalizing a fast nearest neighbor search algorithm, and given a few preliminary experiments. However, much is left to be done, and we want especially to explore further the following questions:

- What sort of data are particularly suited for lazy learning by analogy? We know from the bibliography that linguistic data have been successfully processed with learning by analogy techniques, in fields such as grapheme to phoneme transcription, morphology, translation. We are currently working on experiments on phoneme to grapheme transcription, which can be useful in some special cases in speech recognition (for proper names, for example). We also are interested on other sequential real data, such as biosequences, in which the analogical reasoning technique is (rather unformally) presently already used. The selection of the data and of the supervision are equally important, since both the search of the less dissemblant analogic triple and the labelling process are based on the same concept of analogy.

- What sort of structured data can be processed? Sequences can naturally be extended to ordered trees, in which several generalizations of alignments have already been defined. This could be useful, for example, in extending the nearest neighbor technique in learning prosodic trees for speech synthesis (Blin & Miclet, 2000). We could also imagine that sequences models, like Hidden Markov Models (HMM) could be combined through an analogical construction.

- What sort of algorithms can be devised to let large amount of data be processed by such techniques? We have given a first answer with the FADANA algorithm, and we believe that the quality of the results can be still increased. More experiments remain to be done with this type of algorithm. We have to notice also that not all the properties of analogical dissimilarity have been used so far. We believe that an algorithm with a precomputing and a storage in $\mathcal{O}(m)$ can be devised, and we are currently working on it.

In conclusion, we are confident in the fact that the new notion of analogical dissimilarity and the lazy learning technique that we have associated with it can be extended to more real data, other structures of data and larger problems.

## Acknowledgments

The authors would like to thank the anonymous referrees for their constructive and detailed comments on the first version of this article.

## References

Aamodt, A., & Plaza, E. (1994). Case-based reasoning: Foundational issues, methodological variations, and system approaches. *Artificial Intelligence Communications*, *7*(1), 39–59.

Basu, M., Bunke, H., & Del Bimbo, A. (Eds.). (2005). *Syntactic and Structural Pattern Recognition*, Vol. 27 of *Special Section of IEEE Trans. Pattern Analysis and Machine Intelligence*. IEEE Computer Society.

Bayoudh, S., Miclet, L., & Delhay, A. (2007a). Learning by analogy : a classification rule for binary and nominal data. In Veloso, M. M. (Ed.), *International Joint Conference on Artificial Intelligence*, Vol. 20, pp. 678–683. AAAI Press.

Bayoudh, S., Mouchère, H., Miclet, L., & Anquetil, E. (2007b). Learning a classifier with very few examples: analogy based and knowledge based generation of new examples for character recognition.. In *European Conference on Machine Learning*, Vol. 18. Springer Verlag LNAI 4701.

Bishop, C. (2007). *Pattern Recognition and Machine Learning*. Springer.

Blin, L., & Miclet, L. (2000). Generating synthetic speech prosody with lazy learning in tree structures. In *Proceedings of CoNLL-2000 : 4th Conference on Computational Natural Language Learning*, pp. 87–90, Lisboa, Portugal.

Bunke, H., & Caelli, T. (Eds.). (2004). *Graph Matching in Pattern Recognition and Machine Vision, Special Issue of International Journal of Pattern Recognition and Artificial Intelligence*. World Scientific.

Cano, J., Pérez-Cortes, J., Arlandis, J., & Llobet, R. (2002). Training set expansion in handwritten character recognition.. In *9th Int. Workshop on Structural and Syntactic Pattern Recognition*, pp. 548–556.

Chávez, E., Navarro, G., Baeza-Yates, R., & Marroquín, J.-L. (2001). Searching in metric spaces. *ACM Comput. Surv.*, *33*(3), 273–321.

Cornuéjols, A., & Miclet, L. (2002). *Apprentissage artificiel : concepts et algorithmes*. Eyrolles, Paris.

Daelemans, W. (1996). Abstraction considered harmful: lazy learning of language processing. In den Herik, H. J. V., & Weijters, A. (Eds.), *Proceedings of the sixth Belgian-Dutch Conference on Machine Learning*, pp. 3–12, Maastricht, The Nederlands.

Dastani, M., Indurkhya, B., & Scha, R. (2003). Analogical projection in pattern perception. *Journal of Experimental and Theoretical Artificial Intelligence*, *15*(4).

Delhay, A., & Miclet, L. (2004). Analogical equations in sequences : Definition and resolution.. In *International Colloquium on Grammatical Induction*, pp. 127–138, Athens, Greece.

Dress, A. W. M., Füllen, G., & Perrey, S. (1995). A divide and conquer approach to multiple alignment. In *ISMB*, pp. 107–113.

Edgar, R. (2004). Muscle: a multiple sequence alignment method with reduced time and space complexity. *BMC Bioinformatics*, *5*(1), 113.

Falkenhainer, B., Forbus, K., & Gentner, D. (1989). The structure-mapping engine: Algorithm and examples. *Artificial Intelligence*, *41*, 1–63.

Gentner, D., Holyoak, K. J., & Kokinov, B. (2001). *The analogical mind: Perspectives from cognitive science*. MIT Press.

Gillick, L., & Cox, S. (1989). Some statistical issues in the comparison of speech recognition algorithms.. In *IEEE Conference on Acoustics, Speech and Signal Processing*, pp. 532–535, Glasgow, UK.

Hofstadter, D., & the Fluid Analogies Research Group (1994). *Fluid Concepts and Creative Analogies*. Basic Books, New York.

Holyoak, K. (2005). Analogy. In *The Cambridge Handbook of Thinking and Reasoning*, chap. 6. Cambridge University Press.

Hull, D. (1993). Using statistical testing in the evaluation of retrieval experiments. In *Research and Development in Information Retrieval*, pp. 329–338.

Itkonen, E., & Haukioja, J. (1997). *A rehabilitation of analogy in syntax (and elsewhere)*, pp. 131–177. Peter Lang.

Lepage, Y. (2001). Apparatus and method for producing analogically similar word based on pseudo-distances between words..

Lepage, Y. (2003). *De l'analogie rendant compte de la commutation en linguistique*. Université Joseph Fourier, Grenoble. Habilitation à diriger les recherches.

Micó, L., Oncina, J., & Vidal, E. (1994). A new version of the nearest-neighbour approximating and eliminating search algorithm aesa with linear preprocessing-time and memory requirements. *Pattern Recognition Letters*, *15*, 9–17.

Mitchell, M. (1993). *Analogy-Making as Perception*. MIT Press.

Mitchell, T. (1997). *Machine Learning*. McGraw-Hill.

Mouchère, H., Anquetil, E., & Ragot, N. (2007). Writer style adaptation in on-line handwriting recognizers by a fuzzy mechanism approach: The adapt method. *Int. Journal of Pattern Recognition and Artificial Intelligence*, *21*(1), 99–116.

Needleman, S. B., & Wunsch, C. D. (1970). A general method applicable to the search for similarities in the amino acid sequence of two proteins.. *J Mol Biol*, *48*(3), 443–453.

Newman, D., Hettich, S., Blake, C., & Merz, C. (1998). UCI repository of machine learning databases..

Pirrelli, V., & Yvon, F. (1999). Analogy in the lexicon: a probe into analogy-based machine learning of language. In *Proceedings of the 6th International Symposium on Human Communication*, Santiago de Cuba, Cuba.

Schmid, U., Gust, H., Kühnberger, K.-U., & Burghardt, J. (2003). An algebraic framework for solving proportional and predictive analogies. In F. Schmalhofer, R. Y., & Katz, G. (Eds.), *Proceedings of the European Conference on Cognitive Science (EuroCogSci 2003)*, pp. 295–300, Osnabrück, Germany. Lawrence Erlbaum.

Smith, T. F., & Waterman, M. S. (1981). Identification of common molecular subsequences. *Journal of Molecular Biology*, *147*, 195–197.

Stroppa, N., & Yvon, F. (2004). Analogie dans les séquences : un solveur à états finis. In *TALN 2004*.

Thompson, J. D., Higgins, D. G., & Gibson, T. J. (1994). Improved sensitivity of profile searches through the use of sequence weights and gap excision. *Computer Applications in the Biosciences*, *10*(1), 19–29.

Turney, P. D. (2005). Measuring semantic similarity by latent relational analysis. *Proceedings Nineteenth International Joint Conference on Artificial Intelligence (IJCAI-05)*, *05*, 1136.

Wang, L., & Jiang, T. (1994). On the complexity of multiple sequence alignment. *Journal of Computational Biology*, *1*(4), 337–348.

Witten, I. H., & Frank, E. (2005). *Data Mining: Practical machine learning tools and techniques, 2nd Edition*. Morgan Kaufmann Publishers.

Yvon, F. (1997). Paradigmatic cascades: a linguistically sound model of pronunciation by analogy. In *Proceedings of the 35th annual meeting of the Association for Computational Linguistics (ACL)*, Madrid, Spain.

Yvon, F. (1999). Pronouncing unknown words using multi-dimensional analogies. In *Proceeding of the European conference on Speech Application and Technology (Eurospeech)*, Vol. 1, pp. 199–202, Budapest, Hungary.

Yvon, F., Stroppa, N., Delhay, A., & Miclet, L. (2004). Solving analogical equations on words. Tech. rep. ENST2004D005, École Nationale Supérieure des Télécommunications.